%% file: evo2024.tex
\documentclass[runningheads]{llncs}
\usepackage[T1]{fontenc}
\usepackage{graphicx}
\usepackage{hyperref}
\usepackage{color}

\usepackage{amsmath}
\usepackage{amssymb}
\usepackage{amsfonts}
\usepackage{booktabs}
\newcommand*\sq{\mathbin{\vcenter{\hbox{\rule{1ex}{1ex}}}}}
\usepackage{subfig}
\usepackage[ruled]{algorithm2e}

\newif\ifanonymize
\anonymizefalse

\begin{document}
\title{Evolutionary Algorithms for Optimizing Emergency Exit Placement in Indoor Environments\thanks{\ifanonymize
\else
This work is supported by Spanish Ministry of Science and Innovation under project Bio4Res (PID2021-125184NB-I00 – \url{http://bio4res.lcc.uma.es}) and by Universidad de M\'alaga, Campus de Excelencia Internacional Andaluc\'{\i}a Tech.
\fi}
\thanks{This document is a preprint of Cotta, C., Gallardo, J.E. (2024). Evolutionary Algorithms for Optimizing Emergency Exit Placement in Indoor Environments. In: Smith, S., Correia, J., Cintrano, C. (eds) Applications of Evolutionary Computation. EvoApplications 2024. Lecture Notes in Computer Science, vol 14634. Springer, \url{https://doi.org/10.1007/978-3-031-56852-7_13}}}
\titlerunning{Evolutionary Placement of Emergency Exits in Indoor Environments}
%
\ifanonymize
\else
\author{Carlos Cotta\inst{1,2}\orcidID{0000-0001-8478-7549} \and
Jos\'e E. Gallardo\inst{1,2}\orcidID{0000-0002-0646-2535}}

\institute{
Dept. Lenguajes y Ciencias de la Computación, ETSI Informática, \linebreak
Campus de Teatinos, Universidad de Málaga, 29071 Málaga, Spain
\and
ITIS Software, Universidad de Málaga, Spain\linebreak
\email{\{ccottap,pepeg\}@lcc.uma.es}
}
\authorrunning{C. Cotta \and J. E. Gallardo}
%
\fi
\maketitle              
\begin{abstract}
The problem of finding the optimal placement of emergency exits in an indoor environment to facilitate the rapid and orderly evacuation of crowds is addressed in this work. A cellular-automaton model is used to simulate the behavior of pedestrians in such scenarios, taking into account factors such as the environment, the pedestrians themselves, and the interactions among them.  A metric is proposed to determine how successful or satisfactory an evacuation was. Subsequently, two metaheuristic algorithms, namely an iterated greedy heuristic and an evolutionary algorithm (EA) are proposed to solve the optimization problem. A comparative analysis shows that the proposed EA is able to find effective solutions for different scenarios, and that an island-based version of it outperforms the other two algorithms in terms of solution quality.

\end{abstract}

\input{introduction}
\input{materials}
\input{experimentation}

\input{conclusions}

\ifanonymize
\else
\section*{Acknowledgments}
The authors thank the Supercomputing and Bioinnovation Center (SCBI) of the University of Malaga for their provision 
of computational resources (the supercomputer Picasso \url{http://www.scbi.uma.es}).
\fi

%


\input{evo2024.bbl}
\end{document}

%% file: introduction.tex
\section{Introduction}
\label{sec:introduction}

In the event of an emergency, the rapid and orderly evacuation of crowds from enclosed spaces is essential to minimize casualties and ensure public safety. Needless to say, it can also become a critical challenge requiring meticulous planning at different levels, in order to avoid panic, bottlenecks, and potential harm to people in a potentially chaotic scenario \cite{haghani_optimising_2020}. 
There are different factors that need being taken into account depending on the specificities of each situation (e.g., what the particulars of the environment are, what the typical size and composition of the crowd is, and so on), and the level at which the planning is done (e.g., architectural decisions, signaling, etc.). In this work we are specifically concerned about the placement of emergency exits in the most convenient way to facilitate the efficient evacuation of the crowd.

In order to approach any evacuation optimization problem --such as the one considered here-- and attain safe and efficient evacuation plans, understanding and predicting the behavior of pedestrians is of paramount importance. However, pedestrian evacuation is a complex and dynamic process, influenced by many factors, such as the environment, the pedestrians themselves, and the interactions among them. Therefore, modeling pedestrian evacuation is a challenging task that requires a balance between simplicity and realism. There are different tools that can be used for this purpose, depending on the scope of the simulation. Thus, whereas macroscopic approaches will often consider the crowd as a continuous medium whose flow is to be modeled, e.g., see \cite{bellomo_microscale_2013,golas_continuum_2014}, microscopic models will focus on the pedestrians --the individual components of the crowd-- and model the crowd behavior as an emergent property of the collective behavior of those individual agents. The latter models can be further divided into two major categories, namely models based on social forces (in which pedestrians are particles in a continuous space, subject to different forces resulting from their interaction with the environment and other particles, e.g., \cite{cao_development_2023,li_force-driven_2022}), and cellular-automaton (CA) models (in which the environment is modeled as a discrete grid, and pedestrians transition between these following some predefined rules, e.g.,\cite{shi_dynamic_2019,zheng_simulation_2019}). We refer to \cite{chen21pedestrian,martinez-gil_modeling_2017} for a more in-depth survey of all these approaches. 

We have precisely considered the CA model in this work (see Sect. \ref{sec:automata}). Thus, using this tool for modeling the behavior of a crowd evacuating an indoor environment, we aim to find which would be the most appropriate location for emergency exits. This also entails defining appropriate metrics to assess to which extent an evacuation was successful/satisfactory or not. We do this in Sect. \ref{sec:problem}. Subsequently, we consider different algorithmic approaches to tackle this problem. To be precise, we devise an iterated greedy heuristic and an evolutionary algorithm (EA) for this purpose (see Sect. \ref{sec:algorithms}). We conduct an extensive experimentation to analyze the performance of these algorithms (as well as an island-based version of the EA) in Sect. \ref{sec:experimentation}. Our main aim in this work is to determine the effectiveness of these approaches for this particular optimization setting, as a stepping stone for devising more powerful approaches and tackling more complex evacuation scenarios. We close this work with a critical outlook of the results and an overview of the following steps in this research.

%% file: materials.tex
\input{problem}
\input{automata}

\input{algorithms}

%% file: problem.tex
\section{Problem Statement}
\label{sec:problem}
In order to model the evacuation problem, we need to start by formalizing the indoor space from which the evacuation is attempted. To this end, let ${\cal A}$ be this space, which we will assume to be a rectangular area of width $w$ and height $h$. This rectangular area represents the floor plan of an enclosed space and therefore all its boundaries are assumed to be blocked (i.e., to be non-traversable), except in specific locations which will be denoted as \emph{accesses}. More precisely, we can define an access $\alpha$ as a pair $(p_\alpha, w_\alpha)$, where $p_\alpha$ denotes a point along the perimeter (i.e., a value between 0 and $2(w+h)$, where 0 corresponds to a certain predefined reference point (e.g., the bottom-left corner of ${\cal A}$) of the area at which the access is anchored, and $w_\alpha$ denotes the width of the access along the perimeter, that is, the access extends from $p_\alpha$ to $p_\alpha + w_\alpha$\footnote{Note that since the perimeter is closed, the sum is to be understood as cycling back to 0 when reaching $2(w+h)$.}. Now, within ${\cal A}$ there may be a number of \emph{obstacles}. Each obstacle $o \subseteq {\cal A}$ denotes a non-traversable region (representing real-world objects such as walls or furniture). Therefore, the whole environment can be represented as a tuple $(w, h, A, O)$, where:
\begin{itemize}
	\item $w$ and $h$ are the width and height of ${\cal A}$ respectively.
	\item $A = \{\alpha_1, \dots, \alpha_k\}$ is a collection of accesses.
	\item $O = \{o_1, \dots, o_m\}$ is a collection of obstacles.
\end{itemize} 
This environment is crowded with $n$ pedestrians (they represent the users of said environment, i.e., residents, workers, customers, etc. depending on what it is being modeled) distributed along traversable areas of ${\cal A}$. At time $t=0$, an emergency is declared and the evacuation of the place begins. Let $\mathbb{M}$ be a model that can be used to predict the behavior of pedestrians in this context, and how the evacuation process would then be conducted (cf. Sect. \ref{sec:automata}). Let $\rho_i(t)$ represent the position coordinates of the $i$-th pedestrian at time $t$, and let $T$ be the maximum time up to which the model is simulated. Then, we can split the collection of pedestrians into two sets:
\begin{itemize}
	\item \emph{evacuees} $\xi^+ = \{\ i\ |\  1\leqslant i\leqslant n, \exists t_i\leqslant T: \exists \alpha\in A: \rho_i(t_i)\in \alpha\}$, i.e., all pedestrians $i$ who manage to reach an access before $T$.
	\item \emph{non-evacuees} $\xi^- = \{1,\dots, n\}\setminus \xi^+$, i.e., the pedestrians who could not reach an access before $T$. Given a non-evacuee $i$, we can define $d_i = \min_{\alpha \in A} \lVert \alpha - \rho_i(T)\rVert$, i.e., their distance to the nearest exit at the end of the simulation.
\end{itemize}
In order to quantify the extent to which the evacuation is successful, different metrics could be used. We consider the following hierarchy of objectives:
\begin{enumerate}
\item The first goal is to minimize the number of non-evacuees $|\xi^-|$. This has the highest priority.
\end{enumerate}
The next levels of the hierarchy depend on whether the first goal could be accomplished or not. In the first case ($\xi^-=\emptyset$), we consider:
\begin{enumerate}
\item[2a.] Minimize the time at which the last pedestrian left the area, i.e., minimize $t^* = \max_{1\leqslant i \leqslant n} t_i$.
\item[3a.] Minimize the average time at which pedestrians left the area, i.e., minimize $\bar{t}=\frac{1}{n}\sum_{1\leqslant i \leqslant n} t_i$.
\end{enumerate}
If the evacuation was however not complete, then:
\begin{enumerate}
\item[2b.] Minimize the minimum distance between a non-evacuee and an access, i.e., minimize $d^* = \min_{i\in \xi^-} d_i$.
\item[3b.] Minimize the average distance between non-evacuees and accesses, i.e., minimize $\bar{d} = \frac{1}{n}\sum_{i\in \xi^-} d_i$.
\end{enumerate}
This hierarchy of goals can be combined into a single numerical value by using appropriate weights that ensure that any comparison respects said hierarchy. 
To be precise, let $\sigma({\cal A}, S)$ be a tuple containing the evacuation status of each pedestrian and the corresponding value of $d_i$ or $t_i$ at the end of the simulation,
given that $S=[\rho_1(0),\dots, \rho_n(0)]$ are the initial positions in ${\cal A}$ of the pedestrians at time $t=0$. 
Then, we define:
\begin{equation}
	\begin{array}{rrrl}
	f(\sigma({\cal A},S)) & =  |\xi^-|  + & {[}\xi^-=\emptyset{]} & \left(\frac{1}{T}\max_{1\leqslant i \leqslant n} t_i + \frac{1}{nT^2}\sum_{1\leqslant i \leqslant n} t_i \right) + \\[10pt]
	                                &               + &  {[}\xi^-\neq\emptyset{]} & \left(\frac{1}{D}\min_{i\in \xi^-} d_i + \frac{1}{nD^2}\sum_{i\in \xi^-} d_i \right) 
	\end{array}
\end{equation}
where ${[\cdot]}$ are Iverson brackets, and $D = \sqrt{w^2+h^2}$ is the diagonal of the area.
Now, we can formally define the \textsc{Optimal Evacuation Problem} (OEP) as:
\begin{itemize}
\item[] \textbf{Instance}: a tuple $({\cal A}, \mathbb{S}, k, \omega)$, where
\begin{itemize}
	\item ${\cal A} = (w, h, A, O)$ is the environment.
	\item $\mathbb{S} = \{S_1,\dots, S_l\}$ is a collection of initial configurations of $n$ pedestrians, i.e., for all $1\leqslant i \leqslant l$, $|S_i| = n$.
	\item $k \in \mathbb{N}$ is a non-zero value that indicates the number of emergency exits whose location is sought.
	\item $\omega > 0$ is the width of emergency exits.
\end{itemize}\vspace{1mm}

\item[] \textbf{Solution}: a collection $E=\{e_1, \dots, e_k\} \subset [0, 2(w+h)]$, where each $e_i$ represents the location of an emergency exit and such that 
\begin{equation}
	\psi(E) = \frac{1}{l}\sum_{1\leqslant i\leqslant l} f(\sigma({\cal A}', S_i))
\end{equation}
is minimal, where ${\cal A}'$ is obtained from ${\cal A}$ by adding $\{(e_1,\omega), \dots, (e_k, \omega)\}$ to the existing accesses.
\end{itemize}

Having defined the problem, let us turn our attention to how pedestrian behavior is modeled in next section.

%% file: automata.tex
\newcommand{\staticfield}[2]{\mathcal{S}\mathcal{F}_{#1,#2}}
\newcommand{\shortestPath}[2]{\mathcal{S}\mathcal{P}_{#1,#2}}
\newcommand{\maxShortestPath}{\mathcal{S}\mathcal{P}_\text{max}}
\newcommand{\repulsion}[2]{\mathcal{R}_{#1,#2}}
\newcommand{\reachable}[2]{\mathcal{N}_{#1,#2}}
\newcommand{\reachableCell}[1]{\mathcal{N}_{#1}}
\newcommand{\attraction}[2]{\mathcal{A}_{#1,#2}}
\newcommand{\minAttraction}[2]{\mathcal{A}_{\text{min}_{#1,#2}}}
\newcommand{\desirability}[2]{{\mathcal{D}}_{#1,#2}}

\newcommand{\desirabilityCell}[1]{{\mathcal{D}}_{#1}}

\section{A CA for modeling pedestrian evacuation}
\label{sec:automata}

Cellular automata are simple and powerful tools to simulate complex systems, as they can capture the emergence of global patterns from local interactions. 

In this section, we describe the details of the CA model for pedestrian evacuation. We first define the state of the CA, which is the state of each cell in the environment. We then explain the update procedure, which is the procedure that is used to update the state of the CA on each time step and finally define the transition function, which is the function that determines the probability of a pedestrian moving from one cell to another.

\subsection{State of the CA}
The state of the CA is the state of each cell in the environment (epresented by a regular lattice of square cells). Each cell can be in one of three states:
\begin{itemize}
\item \texttt{empty}: The cell is empty and can be occupied by a pedestrian.
\item \texttt{occupied}: The cell is occupied by a pedestrian.
\item \texttt{obstacle}: The cell is occupied by an obstacle and cannot be occupied by a pedestrian.
\end{itemize}
Some cells in the environment are marked as \emph{exit} cells. These are the cells that the pedestrians want to reach to leave the environment. We assume that pedestrians are rational and will try to find the shortest path to the nearest exit. However, the presence of obstacles and other pedestrians can affect their movement and make them choose alternative paths.
To capture this behavior, we define two concepts for each cell: the \emph{static field} and the \emph{crowd repulsion}. The former is a measure of how close a cell is to an exit. The latter is a measure of how crowded the neighborhood of a cell is, taking into account obstacles and other pedestrians. We use these two concepts to calculate the \emph{desirability} of a cell, which is the probability that a pedestrian will move to that cell.

The static field of a cell is computed using Dijkstra's algorithm, which is a well-known algorithm for finding the shortest path between two nodes in a weighted graph \cite{dijkstra1959note}. We consider the environment as a graph, where nodes are  cells and edges are connections between neighboring cells. The weight of an edge is the geometric distance between the cell centers, if the target cell is not an obstacle and infinity otherwise. Formally, we define the graph as $G = (V, E)$, where $V$ is the set of cells in the environment and $E$ is the set of edges between neighboring cells. The weight function is $w: E \to \mathbb{R}^+$, such that $w(v_i, v_j)$ is the geometric distance between cells $v_i$ and $v_j$, as defined before.
 
Let $\shortestPath{i}{j}$ be the length of the shortest path from cell $(i,j)$ to its nearest exit as computed by Dijkstra's algorithm.
The static field of a cell $(i,j)$ is then defined as:
\begin{equation}
\label{eq:staticField}
\staticfield{i}{j} = 1 - \frac{\shortestPath{i}{j}}{\maxShortestPath}
\end{equation}
where $\maxShortestPath$ is the larger shortest path from any cell in the environment to its nearest exit. This definition makes the static field be in [0,1] and only depend on the relative distance of a cell to its nearest exit. The higher the static field, the closer the cell is to an exit. Notice that, as this field is static, it does not change over time and is only computed once before the simulation.

The crowd repulsion of a cell is computed using the number of reachable cells in its neighborhood. A cell is reachable if it is currently empty and not blocked by an obstacle. For each occupied cell $(i,j)$, let $\reachable{i}{j}$ be the set of reachable cells in its neighborhood. The repulsion of a cell $(i, j)$ is defined as the inverse of one plus the number of reachable cells in this neighborhood:
\begin{equation}
\label{eq:repulsion}
\repulsion{i}{j} = \left(1 + |\reachable{i}{j}|\right)^{-1}
\end{equation}
where $|\cdot|$ denotes the cardinality of a set. This definition makes the repulsion be in (0,1] and depend on how crowded the neighborhood of a cell is. 
The higher the repulsion, the more crowded the neighborhood is.

The desirability of a cell is computed using a combination of the static field and the crowd repulsion. We introduce two parameters to weight the importance of these two factors: the \emph{field attraction bias} $\phi$ and the \emph{crowd repulsion bias} $\zeta$. The field attraction bias reflects how strongly the pedestrians are attracted to the exit cells, while the crowd repulsion reflects how strongly the pedestrians are repelled by the crowded cells. We firstly define the attraction of a cell $(i,j)$ as:
\begin{equation}
\label{eq:attraction}
\attraction{i}{j} = \exp(\phi\cdot\staticfield{i}{j} - \zeta\cdot\repulsion{i}{j})
\end{equation}
In this way, the attraction of a cell is a positive number that increases with the static field and decreases with the crowd repulsion. The higher the attraction, the more desirable the cell is. However, we can make the pedestrian behavior more realistic and adaptive by reducing the reliance on the global knowledge of the environment and by making use of the information available in the local neighborhood. As the attraction of a cell is not enough to capture this behavior, we need to consider instead the \emph{desirability} of a cell, which is defined as the gradient of its attraction. The desirability of a cell reflects how the attraction changes locally by comparing the attraction of the cell with the minimum attraction in its reachable neighborhood. Let $\minAttraction{i}{j}$ denote the minimum attraction in neighborhood of cell $(i,j)$, which is defined as:
\begin{equation}
\label{eq:minAttraction}
\minAttraction{i}{j} = \min_{(k,l) \in \reachable{i}{j}} \attraction{k}{l}
\end{equation}
Then, the desirability of cell $(i,j)$ is defined as:
\begin{equation}
\label{eq:desirability}
\desirability{i}{j} = \epsilon + \attraction{i}{j} - \minAttraction{i}{j}
\end{equation}
where $\epsilon$ is a small number which is added to avoid the desirability being zero ($\epsilon=10^{-5}$ in our implementation). In this way, the desirability of a cell is a positive number that increases with the gradient of the attraction. The higher the desirability, the more likely a pedestrian will move to that cell.
The desirability of a cell is the main input of the local rule that updates the state of each cell on each time step. The local rule is based on a probabilistic transition function that determines the probability of a pedestrian moving from one cell to another.

\subsection{Update Procedure}
The update procedure is the procedure that is used to update the state of the CA on each time step. The procedure is as follows:
\begin{enumerate}
\item We start by marking as empty in the next state the cells that are currently occupied by pedestrians, as they may change depending on their movement.
\item We then mark the cells that are occupied by obstacles in the current state as obstacle in the next state. These cells will not change, as they cannot be occupied by pedestrians.
\item We also mark the exit cells that are occupied by pedestrians in the current state as empty in the next state. This models the evacuation of the pedestrians through the exits. We assume that once a pedestrian reaches an exit, they leave the environment and do not come back.
\item For any other cell that is occupied by a pedestrian in the current state, we compute the desirabilities of reachable neighboring cells. We use the desirability as the probability of a pedestrian moving to that cell and randomly select one neighboring cell according to these probabilities. If the selected cell is not occupied by another pedestrian in the next state, we mark it as occupied by the pedestrian in the next state. This means that the pedestrian moves to that cell. Otherwise, we mark the current cell as occupied by the pedestrian in the next state, i.e., the pedestrian stays in the same cell. This way, we avoid collisions between pedestrians and ensure that each cell can have at most one pedestrian. To ensure fairness among pedestrians, we shuffle the order in which we process occupied cells on each time step.
\end{enumerate}
We consider that each cell in the environment is a square and we denote by $cl$ its side length. We denote the time elapsed for each time step as $\Delta t$. The speed of a pedestrian that moves to a neighboring cell on each time step is then $cl/\Delta t$. We call this the \emph{reference speed} of a pedestrian, and denote it by $v$. However, not all pedestrians may move at the same speed (for instance, some pedestrians may move slower than the reference speed, due to physical or psychological factors). To model this, we introduce for each pedestrian a parameter called \emph{velocity percent} ($v_{p}$), which is a percentage of the reference speed. For example, if $v_{p} = 0.5$ for a pedestrian, their speed would be $0.5v$. We model this by letting $v_{p}$ be the probability of a pedestrian moving to a neighboring cell on each time step so that, on average, their speed would be $v_{p} \cdot v$. 

\subsection{Transition Function}
The transition function is the function that determines the probability of a pedestrian moving from one cell to another. The function is based on the desirability of the neighboring cells. The function is defined as follows:
\begin{equation}
\label{eq:transition}
T(c_i, c_j) = \begin{cases}
P_{i,j} \cdot v_{p}, & \text{if } c_j \text{ is empty or an exit in the current state}\\
0, & \text{otherwise}
\end{cases}
\end{equation}
where $c_i$ and $c_j$ are two neighboring cells, $P_{i,j}$ is the probability of agent in cell $c_i$ to move to cell $c_j$ based on its desirability:
\begin{equation}
	\label{eq:probability}
P_{i,j} = \frac{\desirabilityCell{c_j}}{\sum_{c \in \reachableCell{c_i}}{\desirabilityCell{c}}}
\end{equation}
and $v_{p}$ is the velocity percent of the pedestrian in cell $c_i$. The transition function returns the probability of the pedestrian in cell $c_i$ moving to cell $c_j$ on the next time step. The function is zero if cell $c_j$ is blocked or already occupied by another pedestrian in the current state. The function is also zero if the pedestrian in cell $c_i$ does not move in this time step, which happens with probability $1 - v_{p}$.
The transition function is applied to each occupied cell in the current state, after shuffling the order of the cells. The result of the function and the procedure to avoid collisions is used to update the state of the CA on the next time step. The update procedure is repeated until all the pedestrians have evacuated or a maximum number of time steps (corresponding to time $T$) is reached.

%% file: algorithms.tex
\newcommand*\EA{\textsf{EA\ }}
\newcommand*\iEA{\textsf{iEA\ }}
\newcommand*\greedy{\textsf{greedy\ }}
\newcommand*\EAnsp{\textsf{EA}}
\newcommand*\iEAnsp{\textsf{iEA}}
\newcommand*\greedynsp{\textsf{greedy}}

\section{Algorithms for Emergency Exit Optimization}
\label{sec:algorithms}

As indicated in Sect. \ref{sec:problem}, a solution to problem instance OEP$({\cal A}, \mathbb{S}, k, \omega)$ is a set $E=\{e_1, \dots, e_k\} \subset [0, 2(w+h)]$. The mapping between solutions and their associated objective functions values is not just non-linear, but also not available in closed form, and only computable via a stochastic simulation. Thus, it is complex to design low-level heuristics to construct such solutions. We can however engineer a constructive approach on top of the simulations, based on greedy principles. The core of this approach is shown in Algorithm \ref{alg:greedy}.

\begin{algorithm}[!t]
\caption{Greedy constructive heuristic\label{alg:greedy}}
	\KwData{an instance OEP$({\cal A}, \mathbb{S}, k, \omega)$}
	$E\leftarrow \emptyset$\;
	$\eta \leftarrow \lceil2(w+h)/\omega\rceil$\;
	\For{$i\leftarrow 1$ \KwTo $k$} {
		$p \leftarrow$ rand$(0,2(w+h))$\;
		\emph{best}\,$\leftarrow \infty$\;
		\For{$j\leftarrow 1$ \KwTo $\eta$} {
			\emph{cur}\,$\leftarrow \psi (E\cup\{p\})$\;
			\lIf{cur $<$ best} {\emph{best}\,$\leftarrow$\emph{cur}; $e\leftarrow p$}
			$p\leftarrow p + \omega$\;
			\lIf{$p>2(w+h)$}{$p\leftarrow p - 2(w+h)$}
		}
		$E\leftarrow E \cup \{e\}$\;
	}
	\Return{$E$} 
\end{algorithm}

This procedure starts by picking a random initial point $p$ along the perimeter. Then all points $p, p+\omega, p+2\omega, \dots, p+\eta\omega$ are potential candidates to place an exit, where the addition is assumed to wrap around the length of the perimeter, and $\eta$ is picked so as to ensure that we cover the whole perimeter. For each candidate, we simulate the system with an emergency exit in the corresponding location (in addition to any other exits that might have been considered in previous steps), and keep the one that returns the best value of the objective function. This is repeated as many times as needed (i.e., $k$ times) to construct the solution. Notice that this procedure involves computing the value of the objective function $\eta\cdot k$ times. Also, this is a randomized procedure and therefore can be iterated as many times to desired to obtain different greedy solutions. We will denote this latter iterated procedure as \greedynsp.

As an alternative to this greedy heuristic, we consider an EA approach. This is a real-coded EA in which individuals are vectors of $k$ values in the range $[0, 2(w+h)]$. We can initially generate such vectors by sampling uniformly at random the search space. Notice that we do not introduce any constraint regarding the non-overlap of exits. Having two overlapping exits is equivalent within the simulation to having a single exit of width $2\omega-$overlap. We pose that this is less convenient than having two exits back to back without overlapping, or those two exits strategically placed somewhere else. For this reason, we expect evolution will get rid of those suboptimal solutions without the need of introducing an explicit constraint. As to mutation, we have opted for a Gaussian perturbation of a single exit, whose amplitude is a certain percentage $\gamma$ of its current value, i.e.,
\begin{equation}
	e' \leftarrow e\cdot (1 + \gamma{\cal N}(0,1))
\end{equation}
where ${\cal N}(0,1)$ is a normally distributed random value of mean 0 and variance 1. As usual, the value of the variable will wrap around $[0, 2(w+h)]$. As for recombination, we have opted for a discrete set-based approach, since standard operators for continuous variables require a meaningful matching between homologous variables in the parental solutions which is not possible (or at least non-trivial) in this problem. 
Our recombination algorithm is depicted in Algorithm \ref{alg:set-recombination}. It creates a set of candidate locations from the union of the individuals being recombined, and makes a sequence of random picks without replacement from this candidate set. The resulting operator is therefore transmitting and assorting, but not necessarily respectful \cite{radcliffe94algebra}. Besides these operators, our EA uses binary tournament solution, and elitist generational replacement. We have also considered an island version of this EA \cite{alba02parallelism}, which divides the population into a number of separate demes (arranged following a certain topology -- a bidirectional ring in our case) which evolve in partial isolation, and periodically migrate the best solution to neighboring demes, who accept these in substitution of their current worst solutions. We will denote our EA and our island-based EA as \EA and \iEA respectively. All algorithms are available in our GitHub repository\footnote{\url{Masked-for-blind-review}}.

\begin{algorithm}[!t]
\caption{Set-based recombination\label{alg:set-recombination}}
	\KwData{two sets $E=\{e_1, \dots, e_k\}$ and $E'=\{e'_1, \dots, e'_k\} $}
	$C\leftarrow E \cup E'$;
	$S\leftarrow \emptyset$\;
	\For{$i\leftarrow 1$ \KwTo $k$} {
		$e \leftarrow$ \textsc{pick} $(C)$; \tcp{makes random selection}
		$S \leftarrow S \cup \{e\}$;
		$C \leftarrow C \setminus \{e\}$\;
	}
	\Return{$S$} 
\end{algorithm}  

%% file: experimentation.tex
\section{Experimental Results}
\label{sec:experimentation}

The different algorithms described in the previous section have been put to test on a collection of problem instances with different features. These instances and the remaining experimental parameters are described in Sect. \ref{sec:setup}. Subsequently, the numerical results will be reported and analyzed in Sect. \ref{sec:results}.

\input{setup}
\input{results}

%% file: setup.tex
\subsection{Experimental Setup}
\label{sec:setup}
To evaluate the performance of different algorithms, we have generated several environments that simulate evacuation scenarios. 
Our instance generator discretizes the evacuation area in the same fashion our CA does (see Sect. \ref{sec:automata}), and places obstacles randomly in the domain, avoiding overlaps and ensuring a minimum distance between them. The obstacles are rectangular and their dimensions are randomly generated as follows: the width of the obstacle can be either one or two cells, if the obstacle is vertical, or between one and 25 cells, if the obstacle is horizontal. The height of the obstacle is inversely proportional to the width, and it can be between one and half of the rows of the domain. The orientation of the obstacle is also randomly chosen, with a 50\% probability of being vertical or horizontal. The position of the obstacle is randomly selected, with the condition that the obstacle does not exceed the boundaries of the domain, and that there is a minimum distance of two cells between the obstacle and any other obstacle, so that the agents can always move around them. The purpose of the obstacles is to create a realistic, diverse, and challenging environment for the agents, by obstructing their movement and forcing them to find alternative paths.

We have generated three sets of instances, each containing five environments with different characteristics depending on the number $|O|$ of osbtacles:
\begin{itemize}
	\item \emph{low-density}: $ |O| \in\{20,\dots, 30\}$. These environments have a low density of obstacles, which means that the agents have more space to move and less chances of colliding with them.
	\item \emph{mid-density}: $|O|  \in\{50,\dots, 75\}$. These environments have a medium density of obstacles, which means that the agents have less space to move and more chances of colliding with them, but still have some room for maneuvering and finding alternative paths.	
	\item \emph{high-density}: $|O| \in\{100,\dots, 150\}$. These environments have a high density of obstacles, which means that the agents have very little space to move and very high chances of colliding with them, and they face a lot of congestion and bottlenecks in their movement.
\end{itemize}
In all cases, the width and height are picked from [40, 50] and [20, 30], the side length of the square cells is 0.5m and no exits are initially placed in the environments. Hence, evacuation will only proceed through the emergency exits placed by the optimization algorithms. We consider three setting in this regard, namely $k\in\{3,4,5\}$ exits. The width of the emergency exits is set to $\omega$ = 2m. All the instances are publicly available in our data repository \cite{evacuation_instances_23}. 

For each instance, we have randomly generated 1000 initial pedestrian configurations. 20 are used as training set for the optimization algorithms, and the remaining ones will be used as test set. In every case we have considered 100 pedestrians. Each of them has a reference velocity $v=1.3$m/s, a velocity percent $v_p\in[0.5, 1]$, field attraction bias $\phi \in[1.5, 2]$, and crowd repulsion bias $\zeta \in[0.25, 0.5]$. The simulation is run up to $T=60$s.

Regarding the algorithms, in all cases we consider $maxevals=20000$. The \EA has a population size $\mu=100$, recombination probability $p_X=0.9$, mutation probability equivalent to a mutation rate $1/\ell$ per variable, where $\ell$ is the number of variables, and gaussian mutation amplitude $\gamma=0.05$. As to the \iEAnsp, it considers 4 islands of size $\mu=25$, and migration frequency of 10 generations. For each algorithm, floor plan and number of exits sought, we perform 20 runs.

%% file: results.tex
\subsection{Results}
\label{sec:results}

\begin{table}
\caption{\label{tab:results:training}Results of the algorithms (out of 20 runs) on the training set. Each column depicts the best
(\(x^*\)), median (\(\tilde{x}\)), mean (\(\bar{x}\)) and standard error
of the mean (\(\sigma_{\bar{x}}\)). For each instance, the algorithm
with the best mean is marked with a star (\(\star\)), and the remaining
algorithms are marked with a symbol that denotes whether the differences are
statistically significant at \(\alpha=0.01\)
(\({\sq}\)), \(\alpha=0.05\) (\(\bullet\)), and
\(\alpha=0.1\) (\(\circ\)) according to a Wilcoxon rank sum test \cite{wilcoxon_individual_1945}.}\tabularnewline

\centering
\resizebox{\linewidth}{!}{
\begin{tabular}[t]{lrrrlrrrlrrrl}
\toprule
\multicolumn{1}{c}{ } & \multicolumn{4}{c}{\greedy} & \multicolumn{4}{c}{\EA} & \multicolumn{4}{c}{\iEA} \\
\cmidrule(l{3pt}r{3pt}){2-5} \cmidrule(l{3pt}r{3pt}){6-9} \cmidrule(l{3pt}r{3pt}){10-13}
instance & $x^*$ & $\tilde{x}$ & $\bar{x} \pm \sigma_{\bar{x}}$ &   & $x^*$ & $\tilde{x}$ & $\bar{x} \pm \sigma_{\bar{x}}$ &   & $x^*$ & $\tilde{x}$ & $\bar{x} \pm \sigma_{\bar{x}}$ &  \\
\midrule
low-density-1-3 & 8.109 & 8.410 & 8.493 $\pm$ 0.045 & $\sq$ & 7.111 & 7.111 & 7.171 $\pm$ 0.023 & $\star$ & 7.111 & 7.111 & 7.184 $\pm$ 0.058 & $\phantom{\circ}$\\
low-density-2-3 & 7.511 & 8.436 & 8.662 $\pm$ 0.216 & $\sq$ & 6.417 & 6.467 & 6.52 $\pm$ 0.02 & $\star$ & 6.417 & 6.488 & 6.534 $\pm$ 0.024 & $\phantom{\circ}$\\
low-density-3-3 & 9.462 & 9.584 & 9.606 $\pm$ 0.02 & $\phantom{\circ}$ & 9.462 & 9.660 & 9.687 $\pm$ 0.035 & $\bullet$ & 9.409 & 9.660 & 9.598 $\pm$ 0.03 & $\star$\\
low-density-4-3 & 11.309 & 11.309 & 11.759 $\pm$ 0.115 & $\sq$ & 9.259 & 9.556 & 9.657 $\pm$ 0.069 & $\sq$ & 9.259 & 9.309 & 9.438 $\pm$ 0.072 & $\star$\\
low-density-5-3 & 8.510 & 8.862 & 8.867 $\pm$ 0.035 & $\sq$ & 4.066 & 4.318 & 4.411 $\pm$ 0.052 & $\circ$ & 4.066 & 4.315 & 4.279 $\pm$ 0.03 & $\star$\\
\addlinespace
mid-density-1-3 & 13.104 & 13.482 & 13.435 $\pm$ 0.039 & $\sq$ & 10.709 & 10.709 & 10.894 $\pm$ 0.118 & $\star$ & 10.709 & 11.509 & 11.613 $\pm$ 0.21 & $\bullet$\\
mid-density-2-3 & 15.306 & 15.432 & 15.477 $\pm$ 0.046 & $\sq$ & 14.507 & 14.507 & 14.507 $\pm$ 0 & $\star$ & 14.507 & 14.507 & 14.507 $\pm$ 0 & $\phantom{\circ}$\\
mid-density-3-3 & 24.656 & 25.007 & 24.919 $\pm$ 0.055 & $\sq$ & 15.555 & 15.555 & 15.688 $\pm$ 0.061 & $\star$ & 15.555 & 15.658 & 15.696 $\pm$ 0.057 & $\phantom{\circ}$\\
mid-density-4-3 & 14.758 & 15.006 & 15.016 $\pm$ 0.019 & $\sq$ & 11.704 & 12.058 & 12.198 $\pm$ 0.1 & $\phantom{\circ}$ & 11.704 & 11.757 & 12.089 $\pm$ 0.128 & $\star$\\
mid-density-5-3 & 13.656 & 14.383 & 14.195 $\pm$ 0.107 & $\sq$ & 12.557 & 12.557 & 12.557 $\pm$ 0 & $\star$ & 12.557 & 12.557 & 12.642 $\pm$ 0.075 & $\phantom{\circ}$\\
\addlinespace
high-density-1-3 & 16.909 & 17.804 & 17.736 $\pm$ 0.083 & $\sq$ & 15.005 & 15.005 & 15.206 $\pm$ 0.126 & $\star$ & 15.005 & 15.005 & 15.436 $\pm$ 0.144 & $\phantom{\circ}$\\
high-density-2-3 & 17.556 & 17.607 & 17.813 $\pm$ 0.11 & $\sq$ & 17.556 & 17.556 & 18.034 $\pm$ 0.477 & $\phantom{\circ}$ & 17.556 & 17.556 & 17.556 $\pm$ 0 & $\star$\\
high-density-3-3 & 25.757 & 25.982 & 26.055 $\pm$ 0.072 & $\sq$ & 18.757 & 19.307 & 19.341 $\pm$ 0.097 & $\phantom{\circ}$ & 18.757 & 19.005 & 19.218 $\pm$ 0.111 & $\star$\\
high-density-4-3 & 17.205 & 17.831 & 17.629 $\pm$ 0.084 & $\sq$ & 15.105 & 15.356 & 15.959 $\pm$ 0.216 & $\star$ & 15.105 & 15.306 & 16.069 $\pm$ 0.236 & $\phantom{\circ}$\\
high-density-5-3 & 13.406 & 13.508 & 13.613 $\pm$ 0.056 & $\sq$ & 13.006 & 13.006 & 13.325 $\pm$ 0.095 & $\phantom{\circ}$ & 13.006 & 13.006 & 13.129 $\pm$ 0.038 & $\star$\\
\addlinespace
low-density-1-4 & 1.316 & 1.510 & 1.492 $\pm$ 0.025 & $\star$ & 1.333 & 1.668 & 1.7 $\pm$ 0.06 & $\bullet$ & 1.263 & 1.738 & 1.611 $\pm$ 0.051 & $\phantom{\circ}$\\
low-density-2-4 & 3.015 & 3.369 & 3.359 $\pm$ 0.04 & $\sq$ & 2.472 & 2.690 & 2.802 $\pm$ 0.084 & $\bullet$ & 2.378 & 2.577 & 2.588 $\pm$ 0.031 & $\star$\\
low-density-3-4 & 2.216 & 2.888 & 2.793 $\pm$ 0.076 & $\sq$ & 1.827 & 2.110 & 2.152 $\pm$ 0.039 & $\sq$ & 1.808 & 1.967 & 2.078 $\pm$ 0.131 & $\star$\\
low-density-4-4 & 2.774 & 3.246 & 3.171 $\pm$ 0.041 & $\sq$ & 2.006 & 2.145 & 2.144 $\pm$ 0.024 & $\phantom{\circ}$ & 1.869 & 2.122 & 2.123 $\pm$ 0.034 & $\star$\\
low-density-5-4 & 1.100 & 1.170 & 1.166 $\pm$ 0.009 & $\sq$ & 1.100 & 1.189 & 1.205 $\pm$ 0.023 & $\sq$ & 1.053 & 1.089 & 1.106 $\pm$ 0.013 & $\star$\\
\addlinespace
mid-density-1-4 & 3.515 & 3.963 & 3.926 $\pm$ 0.041 & $\sq$ & 2.867 & 3.342 & 3.367 $\pm$ 0.048 & $\sq$ & 2.961 & 3.190 & 3.213 $\pm$ 0.045 & $\star$\\
mid-density-2-4 & 5.261 & 5.984 & 5.808 $\pm$ 0.078 & $\circ$ & 5.261 & 5.637 & 5.769 $\pm$ 0.118 & $\phantom{\circ}$ & 5.261 & 5.470 & 5.618 $\pm$ 0.106 & $\star$\\
mid-density-3-4 & 6.161 & 6.513 & 6.523 $\pm$ 0.046 & $\sq$ & 5.610 & 6.111 & 6.1 $\pm$ 0.041 & $\sq$ & 5.610 & 5.860 & 5.87 $\pm$ 0.044 & $\star$\\
mid-density-4-4 & 5.011 & 5.188 & 5.221 $\pm$ 0.046 & $\sq$ & 2.929 & 3.120 & 3.156 $\pm$ 0.026 & $\sq$ & 2.666 & 3.062 & 3.005 $\pm$ 0.041 & $\star$\\
mid-density-5-4 & 5.114 & 5.345 & 5.351 $\pm$ 0.03 & $\bullet$ & 4.967 & 5.188 & 5.266 $\pm$ 0.044 & $\phantom{\circ}$ & 4.915 & 5.263 & 5.247 $\pm$ 0.036 & $\star$\\
\addlinespace
high-density-1-4 & 4.611 & 5.090 & 5.151 $\pm$ 0.082 & $\sq$ & 4.110 & 4.487 & 4.564 $\pm$ 0.055 & $\phantom{\circ}$ & 4.110 & 4.440 & 4.484 $\pm$ 0.068 & $\star$\\
high-density-2-4 & 7.507 & 7.663 & 7.742 $\pm$ 0.057 & $\sq$ & 7.112 & 7.509 & 7.492 $\pm$ 0.047 & $\star$ & 7.112 & 7.360 & 8.889 $\pm$ 0.523 & $\phantom{\circ}$\\
high-density-3-4 & 7.009 & 7.259 & 7.229 $\pm$ 0.041 & $\sq$ & 6.859 & 7.209 & 7.189 $\pm$ 0.038 & $\sq$ & 6.712 & 6.985 & 6.98 $\pm$ 0.031 & $\star$\\
high-density-4-4 & 5.206 & 5.487 & 5.614 $\pm$ 0.073 & $\sq$ & 4.711 & 4.944 & 5.098 $\pm$ 0.09 & $\star$ & 4.519 & 4.786 & 5.424 $\pm$ 0.193 & $\phantom{\circ}$\\
high-density-5-4 & 5.513 & 5.863 & 5.898 $\pm$ 0.032 & $\sq$ & 5.014 & 5.816 & 5.787 $\pm$ 0.07 & $\bullet$ & 4.662 & 5.640 & 5.523 $\pm$ 0.083 & $\star$\\
\addlinespace
low-density-1-5 & 0.985 & 1.053 & 1.055 $\pm$ 0.01 & $\sq$ & 0.968 & 1.025 & 1.035 $\pm$ 0.012 & $\bullet$ & 0.948 & 0.979 & 1.005 $\pm$ 0.012 & $\star$\\
low-density-2-5 & 1.241 & 1.455 & 1.423 $\pm$ 0.017 & $\star$ & 1.295 & 1.561 & 1.608 $\pm$ 0.05 & $\sq$ & 1.151 & 1.611 & 1.513 $\pm$ 0.052 & $\phantom{\circ}$\\
low-density-3-5 & 1.417 & 1.691 & 1.707 $\pm$ 0.035 & $\sq$ & 1.419 & 1.608 & 1.609 $\pm$ 0.02 & $\bullet$ & 1.352 & 1.531 & 1.52 $\pm$ 0.024 & $\star$\\
low-density-4-5 & 1.130 & 1.263 & 1.253 $\pm$ 0.019 & $\bullet$ & 1.192 & 1.250 & 1.254 $\pm$ 0.012 & $\sq$ & 1.062 & 1.168 & 1.198 $\pm$ 0.028 & $\star$\\
low-density-5-5 & 0.942 & 0.964 & 0.965 $\pm$ 0.002 & $\sq$ & 0.924 & 0.945 & 0.95 $\pm$ 0.005 & $\circ$ & 0.908 & 0.935 & 0.941 $\pm$ 0.006 & $\star$\\
\addlinespace
mid-density-1-5 & 1.552 & 1.812 & 1.787 $\pm$ 0.025 & $\sq$ & 1.265 & 1.534 & 1.536 $\pm$ 0.029 & $\circ$ & 1.323 & 1.416 & 1.489 $\pm$ 0.04 & $\star$\\
mid-density-2-5 & 3.520 & 3.971 & 3.98 $\pm$ 0.047 & $\bullet$ & 3.318 & 3.820 & 4.067 $\pm$ 0.12 & $\phantom{\circ}$ & 3.272 & 3.744 & 3.838 $\pm$ 0.081 & $\star$\\
mid-density-3-5 & 3.515 & 4.309 & 4.255 $\pm$ 0.057 & $\bullet$ & 3.515 & 4.036 & 4.016 $\pm$ 0.058 & $\phantom{\circ}$ & 3.513 & 3.912 & 3.99 $\pm$ 0.078 & $\star$\\
mid-density-4-5 & 2.064 & 2.380 & 2.355 $\pm$ 0.05 & $\sq$ & 1.298 & 1.505 & 1.522 $\pm$ 0.035 & $\circ$ & 1.201 & 1.366 & 1.433 $\pm$ 0.035 & $\star$\\
mid-density-5-5 & 2.215 & 2.393 & 2.517 $\pm$ 0.063 & $\sq$ & 1.467 & 1.771 & 1.774 $\pm$ 0.032 & $\sq$ & 1.460 & 1.666 & 1.657 $\pm$ 0.022 & $\star$\\
\addlinespace
high-density-1-5 & 2.112 & 2.411 & 2.419 $\pm$ 0.051 & $\bullet$ & 2.012 & 2.385 & 2.504 $\pm$ 0.084 & $\bullet$ & 1.760 & 2.090 & 2.201 $\pm$ 0.078 & $\star$\\
high-density-2-5 & 4.967 & 5.410 & 5.479 $\pm$ 0.062 & $\phantom{\circ}$ & 4.811 & 5.511 & 5.507 $\pm$ 0.093 & $\phantom{\circ}$ & 4.465 & 5.514 & 5.384 $\pm$ 0.101 & $\star$\\
high-density-3-5 & 4.911 & 5.237 & 5.246 $\pm$ 0.039 & $\sq$ & 4.513 & 4.914 & 4.895 $\pm$ 0.045 & $\sq$ & 4.113 & 4.611 & 4.655 $\pm$ 0.068 & $\star$\\
high-density-4-5 & 3.218 & 3.573 & 3.599 $\pm$ 0.044 & $\sq$ & 2.010 & 2.511 & 2.468 $\pm$ 0.069 & $\sq$ & 1.905 & 2.126 & 2.234 $\pm$ 0.072 & $\star$\\
high-density-5-5 & 2.060 & 2.410 & 2.363 $\pm$ 0.047 & $\sq$ & 1.520 & 1.864 & 1.881 $\pm$ 0.042 & $\phantom{\circ}$ & 1.512 & 1.812 & 1.829 $\pm$ 0.034 & $\star$\\
\bottomrule
\end{tabular}}
\end{table}

\begin{figure}[!t]
\centering
\subfloat[\label{subfig:rank:training}]{\includegraphics[width=0.49\textwidth]{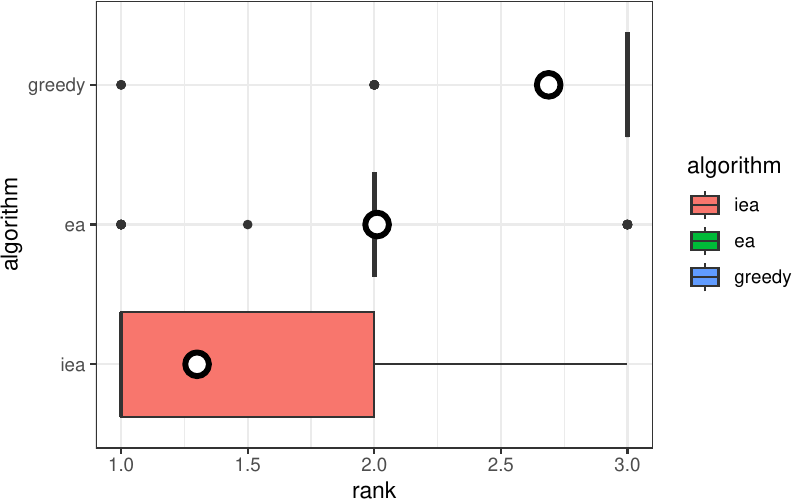}~}
\subfloat[\label{subfig:rank:test}]{~\includegraphics[width=0.49\textwidth]{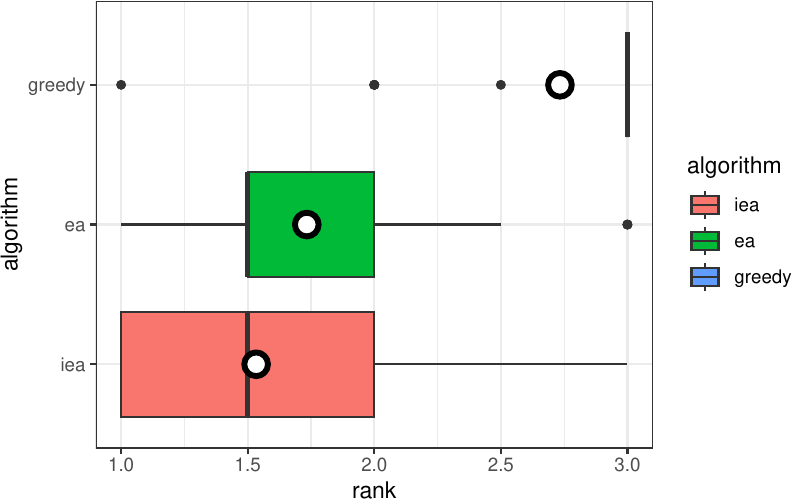}}

\caption{\label{fig:holm}\protect\subref{subfig:rank:training} Rank distribution of the different algorithms on the training set. \protect\subref{subfig:rank:test} Rank
distribution of the best solution of each algorithm on the test set.}
\end{figure}

Table~\ref{tab:results:training} shows the summary of results over the 20 runs of
the algorithms. As it can be seen there is a general superiority of \iEA over all types
of instances and number of exits, and even more clearly for $k\geqslant 4$ exits. This
superiority is not just clear on a head-to-head basis with respect to \EA and \greedy on 
specific instances, but it is also globally significant. To show this, we rank each algorithm 
on each problem instance, and determine de distribution of ranks 
-- see Fig. \ref{subfig:rank:training}. These ranks show statistically significant differences according to 
Quade test \cite{quade_using_1979} (Quade $F$ = 37.351, $p$-value = $1.803e$$-$$12$).
Subsequently, we conduct Holm test with Bonferroni correction \cite{dunn_multiple_1961,holm_simple_1979} using \iEA as control algorithm. The test is passed against both \EA and \greedy
with $p$-value = $7.433e$$-$$4$.

\begin{figure}[!t]
\subfloat[\label{fig:fitness:low}]{\includegraphics[trim={9cm 0 8cm 2cm},clip,width=0.34\textwidth]{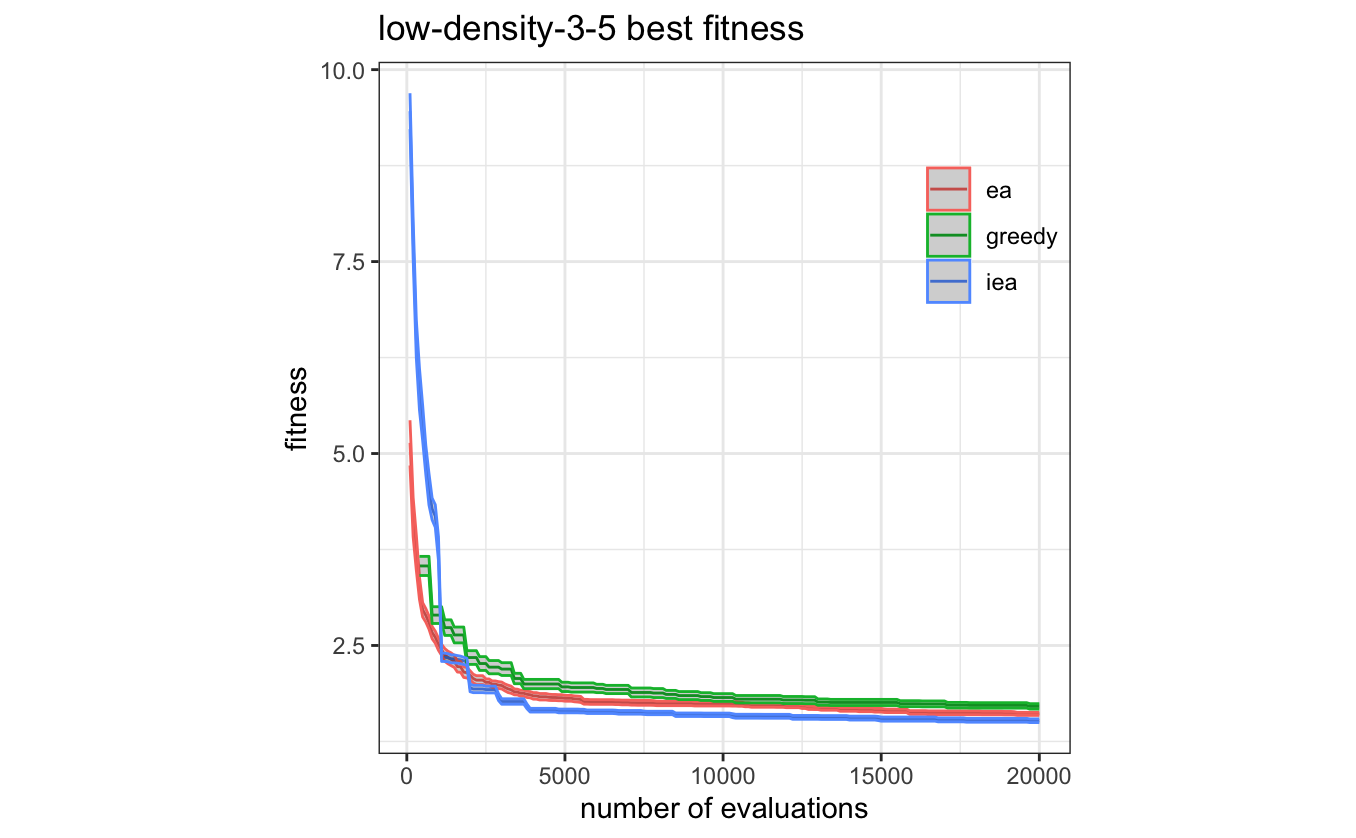}}
\subfloat[\label{fig:fitness:mid}]{\includegraphics[trim={9cm 0 8cm 2cm},clip,width=0.34\textwidth]{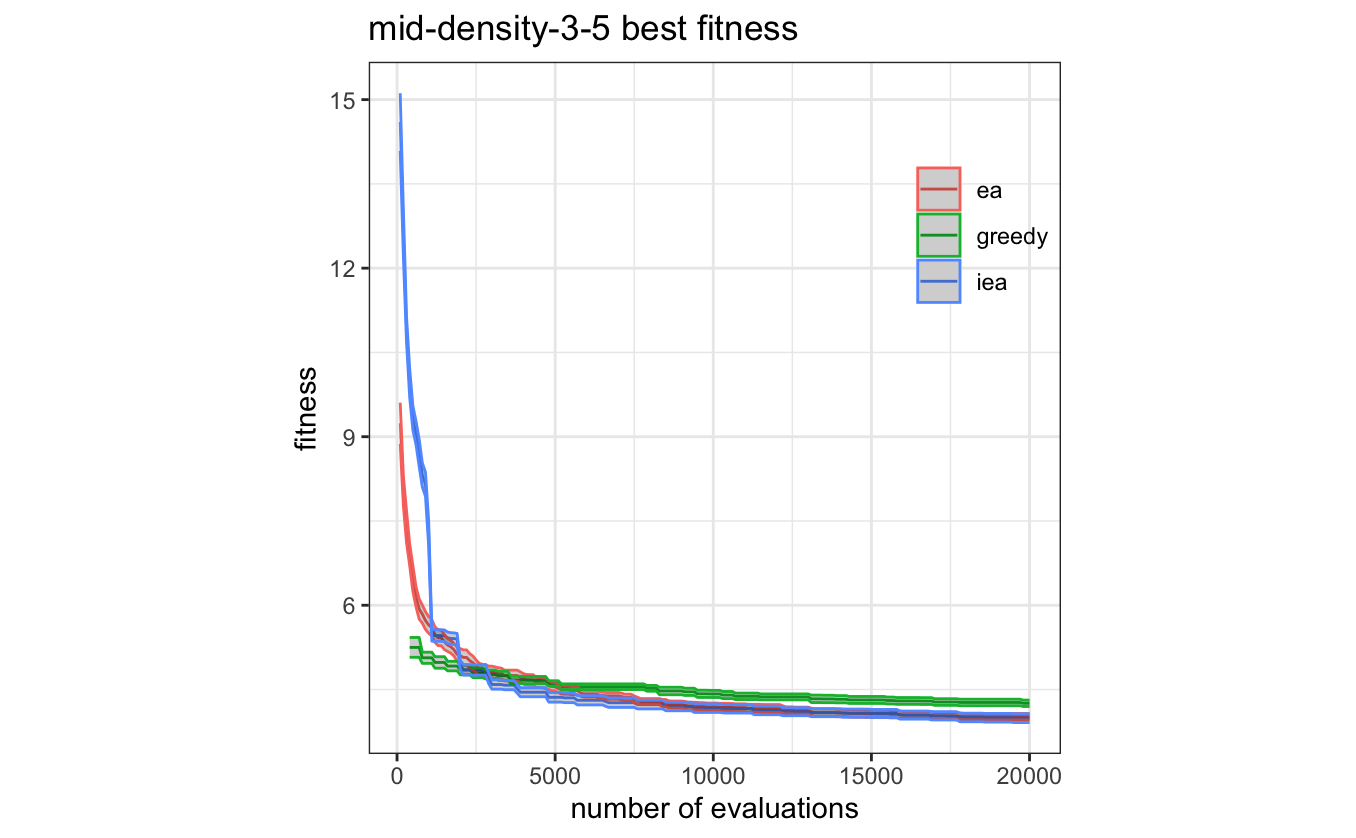}}
\subfloat[\label{fig:fitness:high}]{\includegraphics[trim={9cm 0 8cm 2cm},clip,width=0.34\textwidth]{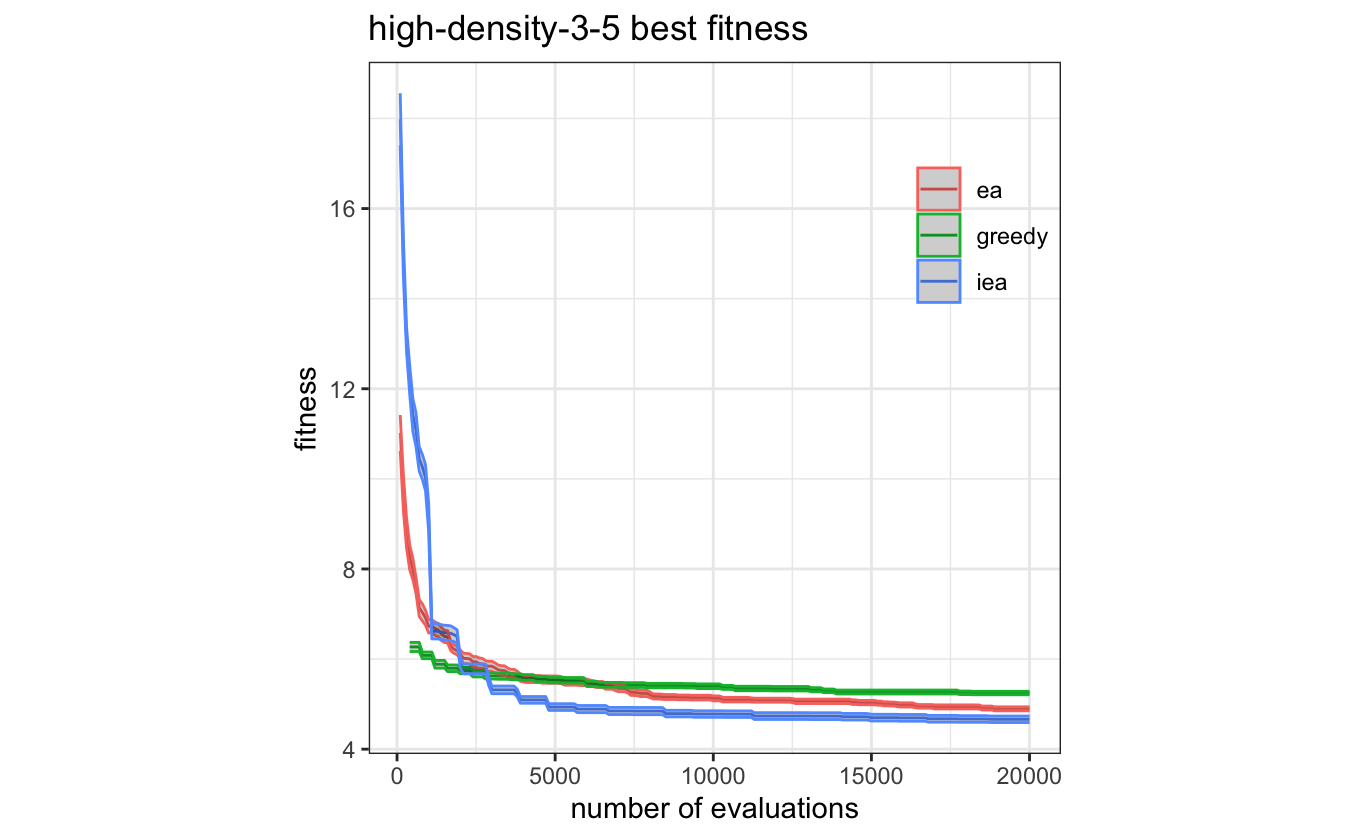}}
\caption{\label{fig:fitness}Evolution of fitness in three of the instances. \protect\subref{fig:fitness:low} low-density  \protect\subref{fig:fitness:mid} mid-density  \protect\subref{fig:fitness:high} high-density}
\end{figure}

These results indicate that the evolutionary search, and in particular the island-based EA, is capable of effectively navigating the search space and finding solutions that perform satisfactorily on the training set. Fig. \ref{fig:fitness} shows an example of the evolution of fitness as a function of the number of solution evaluations for the three algorithms. As it can be seen, \greedy often starts with good quality solutions, typically better than those of \EA and \iEA for a similar computational effort. However, in the long run the evolutionary approaches are capable of outperforming the greedy heuristic. 

\begin{table}
\caption{\label{tab:results:test}Test results of the best solution found by each algorithm during training. 
The meaning of symbols is the same as in Table \protect\ref{tab:results:training}.}\tabularnewline

\centering
\resizebox{\linewidth}{!}{
\begin{tabular}[t]{lrrrlrrrlrrrl}
\toprule
\multicolumn{1}{c}{ } & \multicolumn{4}{c}{\greedy} & \multicolumn{4}{c}{\EA} & \multicolumn{4}{c}{\iEA} \\
\cmidrule(l{3pt}r{3pt}){2-5} \cmidrule(l{3pt}r{3pt}){6-9} \cmidrule(l{3pt}r{3pt}){10-13}
instance & $x^*$ & $\tilde{x}$ & $\bar{x} \pm \sigma_{\bar{x}}$ &   & $x^*$ & $\tilde{x}$ & $\bar{x} \pm \sigma_{\bar{x}}$ &   & $x^*$ & $\tilde{x}$ & $\bar{x} \pm \sigma_{\bar{x}}$ &  \\
\midrule
low-density-1-3 & 2.000 & 9.001 & 8.816 ± 0.093 &$\sq$ & 0.972 & 8.001 & 7.866 ± 0.087 & $\star$ & 0.972 & 8.001 & 7.866 ± 0.087 & $\phantom{\circ}$\\
low-density-2-3 & 2.015 & 8.011 & 8.278 ± 0.084 &$\sq$ & 1.010 & 7.521 & 7.68 ± 0.083 & $\star$ & 1.010 & 7.521 & 7.68 ± 0.083 & $\phantom{\circ}$\\
low-density-3-3 & 3.013 & 11.001 & 10.839 ± 0.098 & $\phantom{\circ}$ & 3.013 & 11.001 & 10.839 ± 0.098 & $\phantom{\circ}$ & 2.009 & 11.001 & 10.673 ± 0.097 & $\star$\\
low-density-4-3 & 4.067 & 12.026 & 12.687 ± 0.11 &$\sq$ & 2.000 & 10.010 & 10.399 ± 0.099 & $\star$ & 2.000 & 10.010 & 10.399 ± 0.099 & $\phantom{\circ}$\\
low-density-5-3 & 1.005 & 9.011 & 8.962 ± 0.091 &$\sq$ & 0.918 & 6.001 & 5.958 ± 0.074 & $\star$ & 0.918 & 6.001 & 5.958 ± 0.074 & $\phantom{\circ}$\\
\addlinespace
mid-density-1-3 & 5.011 & 15.001 & 14.707 ± 0.112 &$\sq$ & 3.010 & 11.010 & 11.34 ± 0.103 & $\star$ & 3.010 & 11.010 & 11.34 ± 0.103 & $\phantom{\circ}$\\
mid-density-2-3 & 6.010 & 16.001 & 15.97 ± 0.11 &$\sq$ & 5.010 & 15.011 & 15.391 ± 0.112 & $\star$ & 5.010 & 15.011 & 15.391 ± 0.112 & $\phantom{\circ}$\\
mid-density-3-3 & 13.010 & 25.002 & 25.242 ± 0.137 &$\sq$ & 8.001 & 17.009 & 17.296 ± 0.119 & $\star$ & 8.001 & 17.009 & 17.296 ± 0.119 & $\phantom{\circ}$\\
mid-density-4-3 & 4.028 & 16.001 & 15.648 ± 0.117 &$\sq$ & 4.019 & 13.001 & 12.831 ± 0.107 & $\star$ & 4.019 & 13.001 & 12.831 ± 0.107 & $\phantom{\circ}$\\
mid-density-5-3 & 4.011 & 15.001 & 15.022 ± 0.112 &$\sq$ & 3.000 & 13.001 & 12.954 ± 0.101 & $\star$ & 3.000 & 13.001 & 12.954 ± 0.101 & $\phantom{\circ}$\\
\addlinespace
high-density-1-3 & 7.010 & 17.001 & 16.852 ± 0.117 &$\sq$ & 6.011 & 15.010 & 15.336 ± 0.108 & $\star$ & 6.011 & 15.010 & 15.336 ± 0.108 & $\phantom{\circ}$\\
high-density-2-3 & 7.011 & 18.001 & 18.038 ± 0.118 & $\star$ & 7.011 & 18.001 & 18.038 ± 0.118 & $\phantom{\circ}$ & 7.011 & 18.001 & 18.038 ± 0.118 & $\phantom{\circ}$\\
high-density-3-3 & 16.001 & 27.002 & 27.097 ± 0.137 &$\sq$ & 8.009 & 20.018 & 20.685 ± 0.127 & $\star$ & 8.009 & 20.018 & 20.685 ± 0.127 & $\phantom{\circ}$\\
high-density-4-3 & 7.010 & 18.510 & 18.585 ± 0.12 &$\sq$ & 7.013 & 17.010 & 17.227 ± 0.115 & $\star$ & 7.013 & 17.010 & 17.227 ± 0.115 & $\phantom{\circ}$\\
high-density-5-3 & 5.011 & 14.011 & 14.365 ± 0.107 & $\phantom{\circ}$ & 5.011 & 14.011 & 14.35 ± 0.109 & $\star$ & 5.011 & 14.011 & 14.35 ± 0.109 & $\phantom{\circ}$\\
\addlinespace
low-density-1-4 & 0.819 & 2.000 & 1.915 ± 0.035 & $\star$ & 0.863 & 2.010 & 2.098 ± 0.039 &$\sq$ & 0.809 & 2.010 & 2.056 ± 0.037 &$\sq$\\
low-density-2-4 & 0.896 & 4.010 & 3.982 ± 0.059 &$\sq$ & 0.852 & 3.011 & 3.039 ± 0.05 & $\star$ & 0.874 & 3.011 & 3.144 ± 0.054 & $\phantom{\circ}$\\
low-density-3-4 & 0.853 & 3.009 & 2.973 ± 0.049 &$\sq$ & 0.863 & 2.029 & 2.669 ± 0.049 & $\star$ & 0.875 & 3.001 & 2.885 ± 0.049 &$\sq$\\
low-density-4-4 & 0.917 & 4.001 & 3.875 ± 0.06 &$\sq$ & 0.809 & 2.042 & 2.714 ± 0.049 & $\star$ & 0.906 & 3.010 & 3.164 ± 0.053 &$\sq$\\
low-density-5-4 & 0.754 & 1.062 & 1.664 ± 0.03 &$\sq$ & 0.754 & 1.021 & 1.422 ± 0.025 & $\phantom{\circ}$ & 0.787 & 1.024 & 1.417 ± 0.024 & $\star$\\
\addlinespace
mid-density-1-4 & 0.917 & 4.020 & 4.429 ± 0.063 & $\circ$ & 0.885 & 4.020 & 4.446 ± 0.064 & $\circ$ & 0.917 & 4.014 & 4.3 ± 0.064 & $\star$\\
mid-density-2-4 & 0.983 & 6.010 & 6.329 ± 0.077 & $\star$ & 0.983 & 6.010 & 6.329 ± 0.077 & $\phantom{\circ}$ & 0.983 & 6.010 & 6.329 ± 0.077 & $\phantom{\circ}$\\
mid-density-3-4 & 2.000 & 7.013 & 7.408 ± 0.082 &$\sq$ & 1.009 & 6.009 & 6.216 ± 0.075 & $\star$ & 1.009 & 6.009 & 6.216 ± 0.075 & $\phantom{\circ}$\\
mid-density-4-4 & 0.994 & 6.010 & 6.166 ± 0.074 &$\sq$ & 0.907 & 4.010 & 4.201 ± 0.062 & $\phantom{\circ}$ & 0.939 & 4.010 & 4.177 ± 0.062 & $\star$\\
mid-density-5-4 & 1.010 & 7.000 & 6.691 ± 0.078 &$\sq$ & 0.994 & 7.001 & 6.904 ± 0.081 &$\sq$ & 0.885 & 6.001 & 5.617 ± 0.07 & $\star$\\
\addlinespace
high-density-1-4 & 0.929 & 5.014 & 5.449 ± 0.07 &$\sq$ & 0.907 & 5.010 & 4.978 ± 0.068 & $\star$ & 0.907 & 5.010 & 4.978 ± 0.068 & $\phantom{\circ}$\\
high-density-2-4 & 1.029 & 8.010 & 8.432 ± 0.088 & $\phantom{\circ}$ & 0.972 & 8.010 & 8.251 ± 0.089 & $\star$ & 0.972 & 8.010 & 8.251 ± 0.089 & $\phantom{\circ}$\\
high-density-3-4 & 0.962 & 8.009 & 8.281 ± 0.088 &$\sq$ & 0.972 & 7.020 & 7.627 ± 0.083 & $\star$ & 1.090 & 8.001 & 8.035 ± 0.087 &$\sq$\\
high-density-4-4 & 1.000 & 6.014 & 6.496 ± 0.078 &$\sq$ & 0.961 & 6.009 & 6.021 ± 0.077 & $\star$ & 1.005 & 6.010 & 6.063 ± 0.074 & $\phantom{\circ}$\\
high-density-5-4 & 1.037 & 7.011 & 7.151 ± 0.079 &$\sq$ & 0.972 & 6.011 & 6.102 ± 0.073 &$\sq$ & 1.000 & 6.001 & 5.851 ± 0.071 & $\star$\\
\addlinespace
low-density-1-5 & 0.775 & 1.022 & 1.482 ± 0.026 &$\sq$ & 0.754 & 1.010 & 1.242 ± 0.019 &$\sq$ & 0.743 & 1.003 & 1.174 ± 0.017 & $\star$\\
low-density-2-5 & 0.743 & 1.041 & 1.57 ± 0.027 & $\star$ & 0.732 & 1.042 & 1.638 ± 0.03 & $\phantom{\circ}$ & 0.786 & 1.042 & 1.63 ± 0.03 & $\phantom{\circ}$\\
low-density-3-5 & 0.852 & 2.012 & 2.217 ± 0.043 &$\sq$ & 0.863 & 2.009 & 2.087 ± 0.039 & $\bullet$ & 0.819 & 2.001 & 1.988 ± 0.038 & $\star$\\
low-density-4-5 & 0.797 & 1.027 & 1.486 ± 0.026 & $\phantom{\circ}$ & 0.775 & 1.021 & 1.475 ± 0.026 & $\star$ & 0.743 & 1.027 & 1.54 ± 0.028 & $\phantom{\circ}$\\
low-density-5-5 & 0.742 & 1.011 & 1.23 ± 0.019 &$\sq$ & 0.655 & 0.970 & 1.046 ± 0.012 & $\star$ & 0.689 & 0.971 & 1.06 ± 0.013 & $\phantom{\circ}$\\
\addlinespace
mid-density-1-5 & 0.808 & 2.010 & 2.119 ± 0.039 &$\sq$ & 0.831 & 2.000 & 1.922 ± 0.036 & $\phantom{\circ}$ & 0.809 & 2.000 & 1.906 ± 0.036 & $\star$\\
mid-density-2-5 & 0.984 & 5.000 & 4.713 ± 0.069 &$\sq$ & 0.896 & 4.010 & 4.231 ± 0.064 & $\star$ & 0.917 & 4.017 & 4.452 ± 0.066 & $\bullet$\\
mid-density-3-5 & 1.000 & 5.009 & 5.243 ± 0.069 &$\sq$ & 1.000 & 5.009 & 5.243 ± 0.069 &$\sq$ & 0.972 & 5.000 & 4.785 ± 0.067 & $\star$\\
mid-density-4-5 & 0.820 & 2.029 & 2.577 ± 0.046 &$\sq$ & 0.797 & 1.044 & 1.703 ± 0.032 & $\star$ & 0.765 & 1.069 & 1.844 ± 0.035 &$\sq$\\
mid-density-5-5 & 0.884 & 3.011 & 3.119 ± 0.052 &$\sq$ & 0.830 & 2.011 & 2.182 ± 0.041 & $\phantom{\circ}$ & 0.830 & 2.011 & 2.142 ± 0.039 & $\star$\\
\addlinespace
high-density-1-5 & 0.863 & 2.022 & 2.514 ± 0.045 &$\sq$ & 0.842 & 2.028 & 2.577 ± 0.045 &$\sq$ & 0.863 & 2.014 & 2.218 ± 0.042 & $\star$\\
high-density-2-5 & 1.010 & 7.000 & 6.7 ± 0.081 &$\sq$ & 0.928 & 6.000 & 5.803 ± 0.074 & $\star$ & 0.950 & 6.001 & 5.943 ± 0.074 & $\phantom{\circ}$\\
high-density-3-5 & 0.994 & 6.009 & 6.148 ± 0.075 &$\sq$ & 0.950 & 6.000 & 5.755 ± 0.073 &$\sq$ & 0.929 & 5.009 & 5.201 ± 0.07 & $\star$\\
high-density-4-5 & 0.885 & 3.036 & 3.581 ± 0.059 &$\sq$ & 0.896 & 3.000 & 2.802 ± 0.048 & $\star$ & 0.852 & 3.000 & 2.819 ± 0.049 & $\phantom{\circ}$\\
high-density-5-5 & 0.830 & 2.021 & 2.436 ± 0.044 & $\phantom{\circ}$ & 0.830 & 2.024 & 2.511 ± 0.045 & $\circ$ & 0.831 & 2.021 & 2.379 ± 0.042 & $\star$\\
\bottomrule
\end{tabular}}
\end{table}

Subsequently, we move to the test phase. To this end, we select the solution that has the best fitness on each instance for each algorithm, and evaluate it on all the test cases. Table \ref{tab:results:test} shows the resulting results. 
Note that \iEA remains superior in general, and both EAs outperform \greedynsp. However, the differences are less marked. 
This is better seen in Fig. \ref{subfig:rank:test}, where the rank distribution of the different algorithms according to the performance of their solution on the test set is shown. Again, these ranks show statistically significant differences 
according to Quade test (Quade $F$ = 50.376, $p$-value = $2.618e$$-$$15$), and \iEA remains the algorithm
with the best mean rank, so it is chosen as control algorithm for Holm test. Now, the test is passed  against \greedy
($p$-value $\approx 0$), but not against \EA ($p$-value = $3.428e$$-$$1$). We believe this may be an indication that the training set is not large enough and therefore \iEA may be overfitting its solutions.

%% file: conclusions.tex
\section{Conclusions}
\label{sec:conclusions}
Optimizing the placement of emergency exits in indoor environments is not just a problem of importance for public safety, but also poses a challenging optimization task. 
We have conducted a comparative analysis of two different optimization approaches, namely an iterated greedy heuristic and an evolutionary algorithm (in two variants,
both panmictic and island-based). This analysis indicates the superiority of the evolutionary approaches, underpinning the need for powerful global optimization techniques in
this context. It also hints at the need of using larger training sets, which of course will have a toll on computational cost. This makes a strong case for directing
effort into solutions of computational nature (such as parallel computing) and solutions of algorithmic nature (e.g., lightweight simulations or surrogate models \cite{jin_surrogate-assisted_2011}).

In addition to the research directions sketched above, it is clear that the evacuation scenario can be enriched with additional layers of complexity. While we have here assumed 
situations of orderly evacuation as an initial base case, we can go on to consider situations in which the cause of the emergency does pose a visible threat (e.g., a rampant fire, 
or ongoing explosions) that might disrupt the evacuation process or the flow of people. Such scenarios may be in need of more sophisticated approaches, and this work has paved the way for 
hybrid approaches that combine greedy components within an evolutionary search engine. Work is in progress in this area. 